\documentclass[letterpaper]{article} 
\usepackage[]{aaai2026}  
\usepackage{times}  
\usepackage{helvet}  
\usepackage{courier}  
\usepackage[hyphens]{url}  
\usepackage{graphicx} 
\usepackage{natbib}  
\usepackage{caption} 
\usepackage{algorithm}
\usepackage{algorithmicx}
\usepackage{algpseudocode}

\usepackage{newfloat}
\usepackage{listings}

\usepackage{tcolorbox}

\DeclareCaptionStyle{ruled}{labelfont=normalfont,labelsep=colon,strut=off} 
\floatstyle{ruled}
\newfloat{listing}{tb}{lst}{}
\floatname{listing}{Listing}
\usepackage{tabularx}
\usepackage{booktabs}
\usepackage{xcolor}  

\title{
From Woofs to Words: \\
Towards Intelligent Robotic Guide Dogs with Verbal Communication
}
\author{
    Yohei Hayamizu\equalcontrib\thanks{Corresponding author: yhayami1@binghamton.edu}, 
    David DeFazio\equalcontrib, 
    Hrudayangam Mehta\equalcontrib, 
    Zainab Altaweel, 
    Jacqueline Choe, \\
    Chao Lin, 
    Jake Juettner, 
    Furui Xiao, 
    Jeremy Blackburn, 
    Shiqi Zhang
}
\affiliations{
    The State University of New York at Binghamton \\
    4400 Vestal Pkwy E, Binghamton, NY 13902 USA \\
}
\usepackage{bibentry}

\begin{document}

\maketitle

\begin{abstract}

Assistive robotics is an important subarea of robotics that focuses on the well-being of people with disabilities. A robotic guide dog is an assistive quadruped robot that helps visually impaired people in obstacle avoidance and navigation. 
Enabling language capabilities for robotic guide dogs goes beyond naively adding an existing dialog system onto a mobile robot. 
The novel challenges include grounding language in the dynamically changing environment and improving spatial awareness for the human handler. 
To address those challenges, we develop a novel dialog system for robotic guide dogs that uses LLMs to verbalize both navigational plans and scenes. 
The goal is to enable verbal communication for collaborative decision-making within the handler-robot team. 
In experiments, we conducted a human study to evaluate different verbalization strategies and a simulation study to assess the efficiency and accuracy in navigation tasks. 

\end{abstract}

\begin{links}
\link{Website}{https://sites.google.com/view/woofs-words}
\end{links}

\section{Introduction}

Guide dogs play a crucial role in the lives of visually impaired individuals, enhancing their independence, confidence, companionship, and mobility~\cite{whitmarsh2005benefits}. 
Unfortunately, breeding and training guide dogs takes multiple years and can be very costly~\cite{tomkins2011behavioral}, and even then, guide dog training centers have less than a 50\% graduation rate~\cite{marks2025acm}. 
At the same time, many people with visual impairments need guide dogs but have difficulties caring for them, e.g., walking miles a day and dog allergies. 
As a result, only a very small portion of the visually impaired use guide dogs, e.g., about 2\% in the U.S.~\cite{marks2025acm} 
and only about 400 guide dogs for more than 10 million visually impaired people in China~\cite{china2023}.

While biological guide dogs are helpful in leading the visually impaired individuals around obstacles, they have fundamental limitations in understanding and communicating with humans. 
Even after systematic training, guide dogs can only respond to very short, predefined verbal cues~\cite{guidingEyes}.
Robotic guide dogs have the potential to be valuable navigation aids for visually impaired individuals~\cite{wang2021navdog, xiao2021robotic, chen2023quadruped, hwang2023system, kim2023transforming, defazio2023seeing, hwang2024towards, people2025synthetic}. 
To the best of our knowledge, there was no existing research focusing on natural language interactions for robotic guide dogs. 

\begin{figure}[t]
\begin{center}
    \includegraphics[scale=0.19]{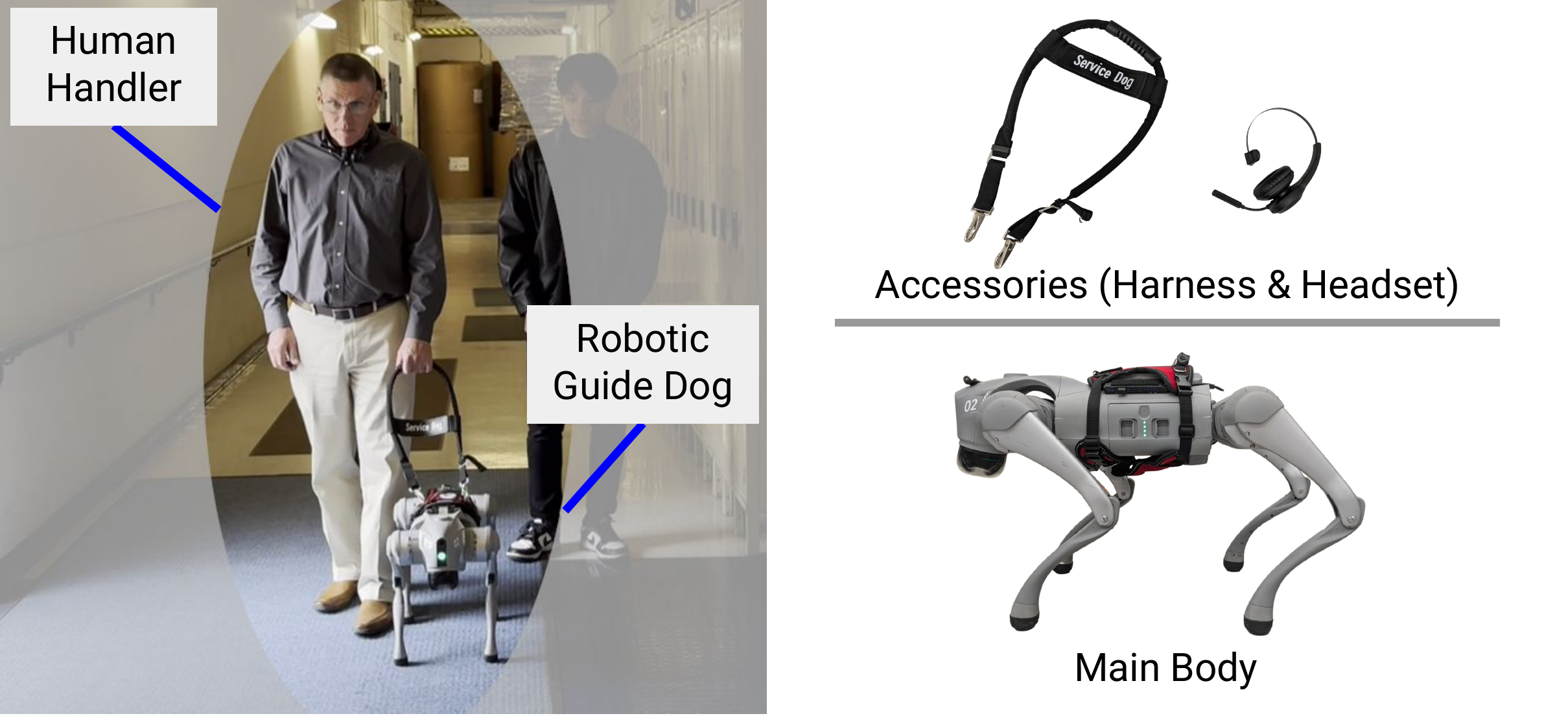}
    \caption{A legally blind person walking with our robotic guide dog system during a participant study. 
    }
    \label{fig:demo_fig}
\end{center}
\end{figure}

The integration of dialog systems into robotic guide dogs introduces unique challenges not present in virtual dialog agents.
First, it is important for the dialog system to ground open-vocabulary language to the current environment, to generate relevant and feasible navigation goals, as well as navigational plans for realizing the goals. 
For instance, when a handler says ``\emph{I am thirsty}'', the dialog system should provide options that are available, e.g., by saying ``\emph{There is a vending machine that serves water bottles and a water fountain on this floor}'', and identify the handler's preference after providing additional information, such as ``\emph{Navigating to the water fountain requires going through a long corridor and entering a kitchen room.}''. 
A second challenge is ensuring the spatial awareness of the visually impaired handler, which is crucial for the handler to make informed decisions about routes and destinations.\footnote{A common misunderstanding is that a visually impaired handler completely follows the biological guide dog in navigation. The reality is that they work as a team, where the handler makes decisions about the routes (e.g., using tactile and echolocation sensing), and the guide dog helps avoid obstacles. }


\begin{figure*}[htp]
\begin{center}
    \includegraphics[scale=0.31]{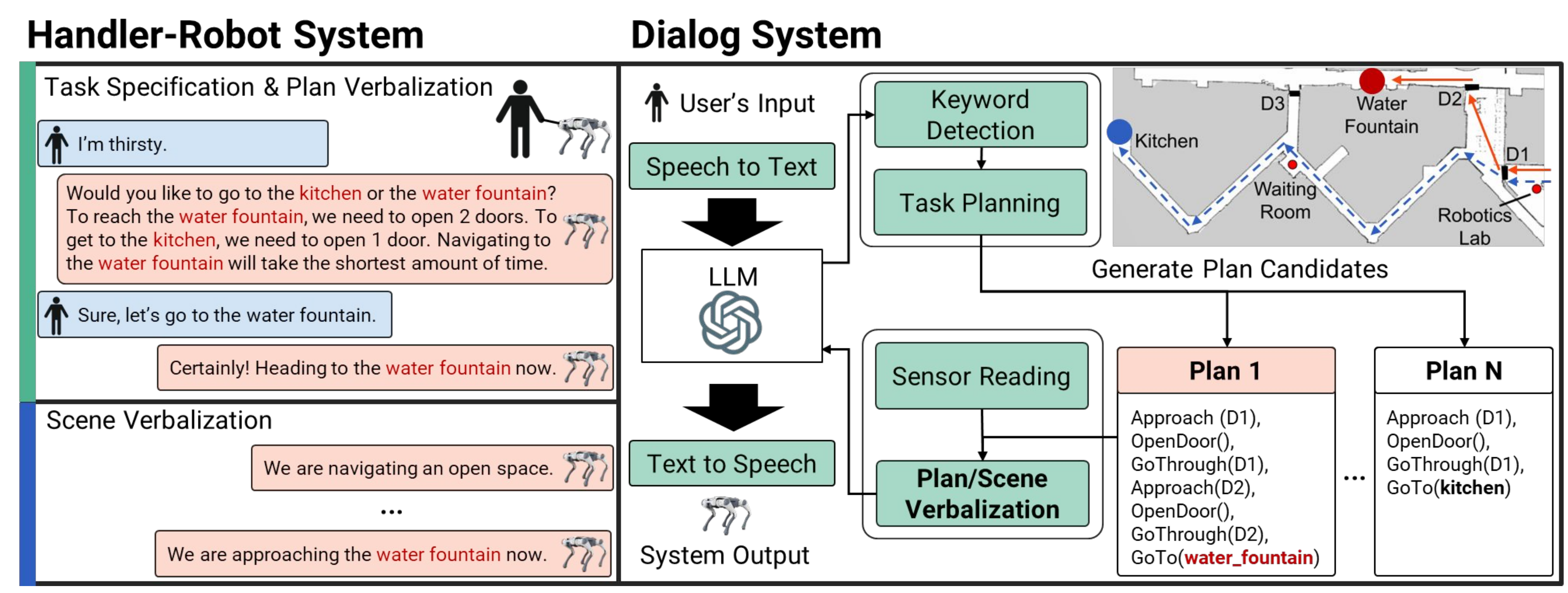}
    \caption{Our system uses human-robot dialog to define a formal service task. An LLM first determines relevant navigable locations, and a task planner generates multiple action sequences (plans) for each candidate. These plans are summarized for the human via \textbf{plan verbalization}, detailing metrics like navigation cost and door openings. After human selection, the robot executes the chosen plan, providing navigation guidance while providing \textbf{scene verbalization} to describe the surroundings.
    }
    \label{fig:system_pic}
\end{center}
\end{figure*}

In this paper, we develop a dialog system that leverages a large language model (LLM)~\cite{achiam2023gpt} for dialog management and extracting navigation tasks from open-vocabulary service requests. 
We integrate task planning into language synthesis to enrich LLM responses with candidate \emph{navigation plans}, which is referred to as \textbf{plan verbalization}. 
After the handler-robot team agrees on a navigation task, the robot will take the sequence of actions generated by the planner, e.g., open door $X$, enter corridor $Y$, and approach room $Z$, to fulfill the service request. 
During navigation, our dialog interface verbalizes scene descriptions to improve the human handler's spatial awareness and facilitate collaborative decision-making, referred to as \textbf{scene verbalization}.
The plan verbalization (before navigation) and scene verbalization (during navigation) capabilities together identify the main novelty of our proposed dialog system for robotic guide dogs. 
We demonstrate the capabilities of our guide dog robot (see Figure~\ref{fig:demo_fig}) through a human study with visually impaired individuals. 
This study, which involves real-world indoor navigation tasks, allowed us to evaluate the system's performance and analyze its potential societal contributions.
To supplement these real-world findings, we also conducted simulation experiments to quantitatively measure the accuracy of our dialog system in identifying navigation tasks from open-vocabulary language.

\section{Related Work}

In this section, we discuss existing robotic guide dog systems, human-robot dialog systems, and non-robotic navigational aid systems for visually impaired people, highlighting how our system differs in each of these categories.

\paragraph{Robotic Guide Dog Systems}have been developed to assist visually impaired individuals in navigating, thanks to reduced hardware costs and recent advances in quadruped locomotion control.
Systems of this type perform human-robot co-navigation tasks, where a human holds a leash to establish a physical connection with a quadruped robot~\cite{cohav2025looks, people2025synthetic}. 
Some of these systems only support unidirectional robot-to-human communication via the forces the human feels from the leash~\cite{wang2021navdog, xiao2021robotic, chen2023quadruped}. 
In those systems, the robot helps the human avoid obstacles at the motion level, and the human completely follows the robot during navigation, assuming both sides know the goal ahead of time. 
Other systems allow for bidirectional communication, such that the human can indicate desired navigation directions via buttons~\cite{hwang2023system}, predefined voice commands~\cite{kim2023transforming, hata2024spot, kim2025understanding}, or tugs on the leash~\cite{defazio2023seeing} during navigation. 
Compared with those robotic guide dog systems, ours further supports human-robot communication via bi-directional, multi-turn dialog. 

\paragraph{Robot Dialog Systems}have greatly improved the interaction capabilities of robots~\cite{tellex2020robots, bisk2020experience,hayamizu2023learning}. 
These systems are optimized for different goals, including mediating the human-robot perception gap~\cite{chai2014collaborative}, learning from dialog experiences~\cite{amiri2019augmenting}, social dialog~\cite{chen2018gunrock}, learning navigation actions from dialog histories~\cite{thomason2020vision}, and understanding ambiguous human instructions~\cite{thomason2015learning}. Earlier works have focused on constructing formal navigation actions from natural language instructions~\cite{chen2011learning} or commands~\cite{matuszek2013learning}. 
While effective for various dialog tasks, none of the above works were designed for the context of guided navigation for visually impaired individuals, where communicating navigation goals, planning to achieve them, and maintaining spatial awareness are critically important. 
Such challenges motivated the development of dialog systems for robotic guide dogs in this paper, where we leverage task planners~\cite{silver2022pddl,liu2023llm} and LLMs~\cite{brohan2023can,stark2023dobby} to facilitate human-robot dialog on navigation tasks in open vocabulary. 

\paragraph{Non-robotic Navigation Aids}were developed to help visually impaired people avoid obstacles, e.g., through belts~\cite{low_vision_run}, dialog systems~\cite{balata2018conversational, katz2012navig, berka2022misalignment, helal2001drishti, chen2020smart}, tele-assistance~\cite{balata2015navigation}, and smart canes that support haptic feedback from the users~\cite{agrawal2022novel}. 
None of those systems were able to ground ambiguous human requests to navigation plans or used a robot for physical guidance, which is believed to be more effective in a recent questionnaire study~\cite{hwang2024towards}. 
This paper showcases a novel dialog system for robotic guide dogs with the capabilities of language synthesis (using a task planner and LLMs) for plan and scene verbalization in handler-robot co-navigation tasks. 

\section{Problem Settings and Methodology}
\label{sec:methodology}

In this section, we present our problem setting and robotic guide dog system as summarized in Figure~\ref{fig:system_pic}. 

\subsection{Problem Settings and Assumptions}
In this paper, we assume the robot has full knowledge about the domain, including a map of the environment associated with semantics information (e.g., labels and coordinates of kitchens and conference rooms). 
We also assume access to a stable quadruped locomotion controller that can accurately track linear and angular velocity commands.
On top of the locomotion controller, the robot is equipped with a global path planner for 2D trajectory planning, and a local path planner that minimizes the deviation from the global path while avoiding obstacles. 
Those software packages are widely used and publicly available in the Robot Operating System~(ROS) community~\cite{koubaa2017robot}. 
We do not discuss scenarios where the human and robot together explore a new environment, the human helps the robot dog recover from falls, or the robot disobeys human commands, which occurs on biological guide dogs. 

Our goal is to develop a robotic guide dog system that fulfills a handler's request by recommending and executing navigation actions based on the service request in spoken language.
This involves developing a dialog system that can map the service request to known locations in the environment, as well as generating an action sequence leading the handler-robot team to the goal location. 

\subsection{Our Dialog System for Robotic Guide Dogs}

Our dialog system combines the flexible natural language understanding capabilities of an LLM with the long-horizon reasoning of task planners.
This integration allows the system to engage in dialogue with users to specify navigation tasks and to deliver contextual, task-relevant information throughout execution. 
A key contribution of our work is the ability to verbalize robotic guide dog's outputs, e.g., scene understanding and decision-making process, into clear, accessible language. 
This verbal feedback equips visually impaired users with the critical information needed to make confident, informed decisions about their own navigation.

We first introduce how our system specifies a task and verbalizes plans based on the task, then describe the scene verbalization process during navigation. 
The control loops are shown in Procedure~\ref{alg:guide_dog_system}.

\floatname{algorithm}{Procedure}
\begin{algorithm}[t!]
\caption{Robotic Guide Dog Dialogue and Navigation}
\label{alg:guide_dog_system}
\small
\begin{algorithmic}[1]

\Statex \Comment{Dialogue for Task Specification \ \ \ \qquad \qquad \qquad \qquad \qquad \qquad \qquad }
\State Greet the guide dog handler to initiate the conversation 
\While{navigational plan is not finalized}
    \State Receive the handler's service request
    \State Convert the service request to Task $T$ in a formal language
    \State Enter $T$ into a task planner and compute task plans $\textbf{P}$
    \For{each plan $P \in \textbf{P}$}
        \State Verbalize plan $P$ using an LLM \Comment{Plan Verbalization}
        \State \textbf{if} User confirms a plan \textbf{then} Break the while loop
    \EndFor
    \State Request the user to rephrase the service request
\EndWhile
\Statex
\Statex \Comment{Plan Execution (concurrent): Thread 1 on navigation \qquad \qquad \qquad }
\For{each navigation action $a$ in plan $P$}
    \State Execute action $a$
\EndFor
\State Return: arrival at the final destination
\Statex
\Statex \Comment{Plan Execution (concurrent): Thread 2 on dialog \qquad \qquad \qquad \qquad }
\While{a scene change is detected or there is a long silence}
\State Collect surrounding objects $O$ using vision and world map
\State Verbalize $O$ with spatial info \Comment{Scene Verbalization}
\EndWhile
\end{algorithmic}
\end{algorithm}

\subsubsection{Plan Verbalization for Task Specification:}

Navigation begins when the user issues a request to task $T$ in natural language through a headset microphone. 
The speech is converted to text using a speech-to-text model~\cite{vosk} and fed to our LLM agent. 
The agent is set with a system prompt (Figure.~\ref{fig:llm_prompt}) that defines its role as a guide dog, lists valid navigation locations, and specifies the expected output formats.

If the user's command is ambiguous (e.g., ``I'm thirsty''), the LLM clarifies the intent through dialogue.
Instead of merely listing potential destinations (e.g., ``kitchen'' and ``water fountain''), our system performs a unique process we call Plan Verbalization. 
For each potential destination, the system runs a symbolic planner (detailed in Section~\ref{sec:implementation_details}) in the background. 
The planner computes an optimal action sequence and extracts information about physical constraints, such as the total cost (i.e., travel time) and the number of doors that must be opened.

The LLM then incorporates this information into its response, generating an informative prompt like, ``\textit{We can go to the kitchen or the water fountain. The kitchen requires opening one door and will take about three minutes. The water fountain has no doors and will take about one minute. Where would you like to go?}'' 
By verbalizing plan details, we enable the user to make an informed decision based on their preferences and circumstances. 
Since LLMs often struggle with spatial reasoning~\cite{cohn2023dialectical}, our dialog system employs a task planner to compute plans for achieving navigation goals by reasoning about both real-world constraints and the robot's action primitives. 
Once the user confirms their choice, the LLM generates a final formal task and a conversational statement to naturally conclude the dialogue (e.g., ``\textit{Okay, I will guide you to the kitchen.}''), ending the dialogue phase.

\begin{figure}[t!]
\centering
\begin{tcolorbox}[
standard jigsaw,
title=Robot Guide Dog Command Protocol,
opacityback=0,
]
\small

You are a robot guide dog assisting a visually impaired person in navigating an indoor environment. The human is holding a rigid connection attached to you. Your task is to navigate to the requested locations based on the human's commands.
When the human makes a request, you must engage in communication to clarify their intentions. Ask clarification questions until you are confident about the request.

\textbf{Command Syntax:}\

There is one command type: \texttt{goto <parameter>} 

\textbf{Valid Parameters:}\
\begin{tabular}{ll}
\textbullet~robotics lab & \textbullet~staircase \\
\textbullet~elevator & \textbullet~conference room \\
\textbullet~bathroom & \textbullet~waiting room \\
\textbullet~water fountain & \textbullet~kitchen \\
\end{tabular}

\textbf{Handling Requests:}\
\begin{itemize}
\item[\textbullet] \textbf{Unavailable Locations:} Do not navigate to locations not on the list. If an unavailable location is requested, you must state this in your clarification question.
\item[\textbullet] \textbf{Ambiguous Requests:} If a request could refer to multiple valid locations, you must suggest each of them in your clarification question.
\end{itemize}

\textbf{Required Output Format:}\
After determining the correct action, provide your output in one of the two formats below.

\textit{To ask for more information:}

\texttt{CLARIFICATION QUESTION: <question>} 

\textit{To execute a command:}

\texttt{COMMAND <type> <parameter> } \

\texttt{conversational statement} 

\end{tcolorbox}
\caption{LLM prompt defining the role of our robot guide dog dialog system. Given possibly ambiguous human service tasks in natural language, the LLM must conduct a dialog and select the location that best satisfies the task. 
}
\label{fig:llm_prompt}
\end{figure}

\subsubsection{Scene Verbalization for Spatial Awareness:}

Once a destination is confirmed and navigation begins, the system continues to provide information to the user through Scene Verbalization \emph{during} navigation. 
This is triggered by major scene changes or a long silence in the dialogue (see Procedure~\ref{alg:guide_dog_system}), such as when the robot's position crosses the boundary of a semantically labeled region on the map (e.g., moving from a ``corridor'' to a ``kitchen'').

When the robot enters a new area, a message describing the current scene is played through a speaker, e.g., ``\emph{We are navigating in a long corridor... We are approaching an office door... We just arrived at an office door.}''
This allows the visually impaired user to maintain awareness of their surroundings during navigation.

While our current implementation uses a simple strategy, we position it as a first step, drawing on the rich literature on verbalization in human-robot co-navigation~\cite{rosenthal2016verbalization}. 
We focus on evaluating its effectiveness in a human study in this paper, and plan to investigate more advanced methods, such as learning verbalization policies, in future work.

\subsection{Implementation Details}
\label{sec:implementation_details}

\subsubsection{Safeguard for Dialog Management:}
Due to the uncertainty of LLM responses and the lack of spatial awareness, we employ a language synthesis module with a safeguard that post-processes the authentic LLM responses before delivering them to the human user. 
Specifically, during the plan verbalization phase, if the LLM does not respond in the expected format (e.g., no ``\textsc{clarification question}'' or ``\textsc{command}'' indicators in the response), or if it fails to suggest any valid locations as determined by keyword detection, then we replace the LLM response with the message: 

\begin{quote}
\textit{``I can only assist with navigation requests of nearby locations that I know about. Would you like help navigating to somewhere nearby?''}
\end{quote}

\subsubsection{Plan Generation for Navigation:}
To complete navigation tasks specified by our dialog system, we generate navigation plans via an Answer Set Programming (ASP) planner~\cite{lifschitz2019answer}. 
We leverage planners designed for navigation tasks~\cite{hayamizu2021guiding} to determine high-level action sequences. We consider the set of actions:

\begin{lstlisting}[breaklines]
  approach(door_id), 
  gothrough(door_id), 
  opendoor(door_id), 
  goto(location) 
\end{lstlisting}

Arriving at the destination specified by the dialog system involves generating a sequence of actions while minimizing their total cost. 
Action costs between locations are considered and measured based on navigation distances, along with an additional constant cost added for door opening actions. 
Rules in the ASP encode the necessary domain knowledge for navigation. 
For example, the following rule states that P1 can be accessed by P2 via door D, if P1 and P2 are unique locations, D is a door, and P1 and P2 both have door D.

\begin{lstlisting}[breaklines]
dooracc(P1,D,P2) :- 
        hasdoor(P1,D), hasdoor(P2,D), 
        P1 != P2, door(D), 
        location(P1), location(P2).
\end{lstlisting}

LLM-based task planning methods are increasingly mature~\cite{kannan2024smart, rana2023sayplan, liu2023llm, zhang2025llm}, and robotic guide dog practitioners can feel free to implement the task planner using LLMs instead. 
Next, we present a human study for evaluating the effectiveness of our dialog system in collaborative settings between a human handler and a robotic guide dog (Section~\ref{sec:human_study_eval}), followed by an extensive evaluation in simulation focusing on accuracy and efficiency (Section~\ref{sec:simulation_eval}). 

\section{Real-World Effectiveness Evaluation: Human Study}
\label{sec:human_study_eval}

In this section, we detail the human study, which was designed to test the following primary hypothesis: Combining scene and plan verbalization improves user satisfaction and perceived effectiveness for robot-assisted navigation.\footnote{This section focuses on presenting the results of our robotic guide dog system under \emph{controlled autonomy}. The system under \emph{full autonomy} is demonstrated on the project webpage.}

\subsection{Participants}
Table~\ref{tab:participant_demographics_compact} shows the demographics of the participants.
Seven legally blind individuals (5 Male, 2 Female; age range 40--68) were recruited for the study.\footnote{A person is considered legally blind if they meet one of the following criteria in their better-seeing eye, even with the best conventional correction: 1) Visual acuity of 20/200 or less; 2) Visual field of 20 degrees or less. 
}
One participant who completed the study self-identified as not meeting the vision requirement and was excluded from the final analysis. 
Two of the participants had prior experience with a guide dog. 

\begin{table}[ht]
\centering
\small 
\begin{tabular}{lccccc}
\toprule
\textbf{Age} & \textbf{Gender} & \textbf{Impairment} & \textbf{Guide Dog} \\
& & \textbf{Duration (yrs)} & \textbf{Experience} \\
\midrule
68 & Female & 68 & None \\
68 & Male  & 60 & None \\
59 & Male  & 51 & 8 yrs, 2 dogs \\
40 & Male & 40 & None \\
64 & Male & 35 & 18 yrs, 3 dogs \\
57 & Male & 20 & None \\
52 & Female & 15 & None \\
\bottomrule
\end{tabular}
\caption{Participant Demographics.}
\label{tab:participant_demographics_compact}
\end{table}

\subsection{Experimental Setup}

The experiment took place in a large, many-room office environment. 
The task involved navigating a route from an entrance to a conference room. 
We used a Unitree Go2 as a quadruped robot platform and integrated it with our dialog system. 
For safety reasons during the human study, the robot's movements were controlled by an expert operator in a Wizard of Oz setup~\cite{dahlback1993wizard}. 
This approach ensures safety and does not affect the evaluation of the effectiveness of the dialog strategies.


After signing a consent form, participants were briefed on the task. 
They would request to go to a conference room, and the robotic guide dog would guide them under three distinct verbalization conditions in a within-subjects design:
\begin{itemize}
    \item \textbf{Minimum Verbalization:} A control condition with language-based task specification before navigation and then no verbal interaction during navigation. 
    \item \textbf{Scene Verbalization:} The robot only verbalized information about the immediate surroundings (e.g., obstacles and environmental context).
    \item \textbf{Ours (Scene + Plan):} The robot provided both scene and plan verbalization.
\end{itemize}

Given the age and the energy span of our participants, we decided to keep the trials and questionnaire within 30 minutes, and evaluate no more than three methods. 
After completing the task under each condition, participants filled out a 7-item questionnaire shown in Table~\ref{tab:list_of_questions}.
Q1 through Q6 were rated on a 5-point Likert scale, and Q7 was an open-ended feedback question. 
Participants self-administered the questionnaires after task completion, while we provided support by reading the instructions and questions. 

\begin{table}[t]
    \centering
    \small
    \begin{tabularx}{\linewidth}{p{1.9cm}|X}
    \toprule
    \centering
    Q1 \qquad \quad Utility & The natural language interface is a useful component of the system. \\
    \hline
    \centering
    Q2 \qquad \quad Helpfulness & The robot is helpful for visually impaired navigation when compared to having no tools for visually impaired navigation. \\
    \hline
    \centering
    Q3 \qquad \quad Helpfulness & The robot is helpful for visually impaired navigation when compared to having other tools for visually impaired navigation. \\
    \hline
    \centering
    Q4 \qquad \quad Safety & The robot is capable of safely guiding me to a navigation goal. \\
    \hline
    \centering
    Q5 \qquad \quad \quad Ease of comm. & I can easily communicate with the robot. \\
    \hline
    \centering
    Q6 \qquad \quad Preference & Given the same navigation task we performed today I prefer a guide dog to guide me instead of the robot. \\
    \hline
    \centering
    Q7 \qquad \quad Feedback & If you have any questions, comments, or concerns please leave them here.\\
    \bottomrule
    \end{tabularx}
    \caption{The list of questions in the questionnaire. The items assessed interface utility, helpfulness, safety, ease of communication, and preference relative to a real guide dog.}
    \label{tab:list_of_questions}
\end{table}

\begin{table}[t]
\centering
\small
\begin{tabular}{lcccccc}
\toprule
\textbf{Condition} & \textbf{Q1} & \textbf{Q2} & \textbf{Q3} & \textbf{Q4} & \textbf{Q5} & \textbf{Q6} \\
\midrule
Mini. Verb.             & 4.33          & 4.67          & 3.50          & \textbf{4.00} & 4.33          & 3.17 \\
Scene Verb.           & 4.67          & \textbf{4.83} & 4.00          & \textbf{4.00} & 4.00          & 3.17 \\
Ours & \textbf{4.83} & \textbf{4.83} & \textbf{4.33} & 3.83          & \textbf{4.50} & \textbf{3.67} \\
\bottomrule
\end{tabular}
\caption{Average survey ratings for each verbalization condition. Ratings are on a 5-point Likert scale (1: Strongly Disagree, 5: Strongly Agree).}
\label{tab:survey_results}
\end{table}

\subsection{Results}

\begin{figure*}[ht]
\begin{center}
    \includegraphics[scale=0.65]{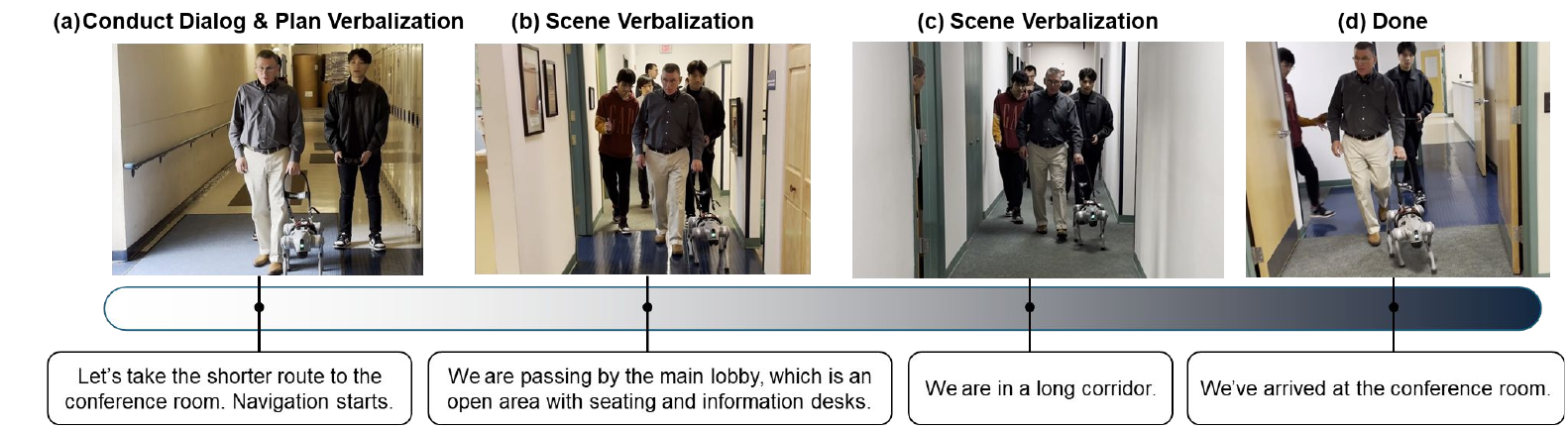}
    \caption{An illustrative example of our robotic guide dog assisting a legally blind participant to navigate to a conference room. 
    The sequence shows:
    (a) The robot verbalizes the generated navigation plans for a user request, and starts navigation after the user chooses one of the plans.
    (b, c) During navigation, the system provides real-time scene verbalization, such as passing a lobby and entering a corridor.
    (d) The robot announces that they have arrived at the destination.
    }
    \label{fig:big_demo_fig}
\end{center}
\end{figure*}

Figure~\ref{fig:big_demo_fig} presents an instance of the experiment trials in the human study. 
Table~\ref{tab:survey_results} shows the average ratings for each condition. 
Our proposed method, \textbf{Ours (Scene + Plan)}, received the highest ratings on natural language interface utility (Q1). 
The result indicates the robot's strong performance in helpfulness (Q2 and Q3).
Furthermore, our method outperformed in ease of communication (Q5) and preferability over real guide dogs (Q6). 
The overall result suggests that providing both scene and plan information leads to higher effectiveness for navigation and a user-friendly approach. 

Our method scored slightly lower on perceived safety (Q4: 3.83) than the other conditions. 
Qualitative feedback (Q7) implies that the reason might be due to initial unfamiliarity with walking with the robot. 
The human study confirms the real-world value of our proposed natural language interface for robotic guide dogs. 


\begin{table}[ht]
\centering
\small
\begin{tabularx}{\linewidth}{ |l|X| }
 \hline
 Category & Locations \\ 
 \hline
 Places to eat & food court, cafeteria, restaurant, dining hall, shawarma joint, cafe, vending machine, food truck, fast food restaurant \\ 
 \hline
 Restroom & restroom, bathroom, the loo \\ 
 \hline
 Change Floor & elevator/stairs, elevator/escalator, elevator \\ 
 \hline
 Cut Hair & barber shop, salon \\ 
 \hline
\end{tabularx}
\caption{Samples in our collected service task dataset: locations in each category considered functionally equivalent.}
\label{tab:functionally_equivalent}
\end{table}

\section{System Performance Evaluation: Simulation Analysis}
\label{sec:simulation_eval}

While the human study demonstrated the real-world usability of our approach, we conducted a large-scale simulation analysis to test the underlying system's performance. 
This analysis tests three hypotheses: our system can accurately and efficiently resolve ambiguous service requests (\textbf{Hypothesis I}), is robust to noisy speech recognition (\textbf{Hypothesis II}), and can leverage a navigation planner to reduce travel costs (\textbf{Hypothesis III}).
Inspired by recent work using LLMs for data synthesis~\cite{chen2023places, tan2024large}, we simulated a visually impaired user with GPT-4~\cite{achiam2023gpt}. 
To ground our simulation in reality, the user's goals were based on a library of navigation-related service requests collected from human participants, as described next.

\subsection{Simulation Setup}

\paragraph{Service Task Library} 

To create a realistic task library, we surveyed 16 university students (ages 19-30). 
Participants were asked to imagine being a visually impaired person who needs help navigating and provide five \texttt{<location, purpose>} pairs for potential navigation requests, such as (\textit{``water fountain''}, \textit{``I want something to drink''}). 
After filtering three overly vague entries, we obtained 77 unique samples. We normalized location names and grouped functionally equivalent locations (e.g., ``cafeteria,'' ``dining hall''), as shown in Table~\ref{tab:functionally_equivalent}. 
The unique locations can be inserted in the system prompt as the set of valid parameters in Figure~\ref{fig:llm_prompt}.
We report accuracy and efficiency for evaluation.

\begin{figure}[t!]
\begin{center}
    \includegraphics[scale=0.30]{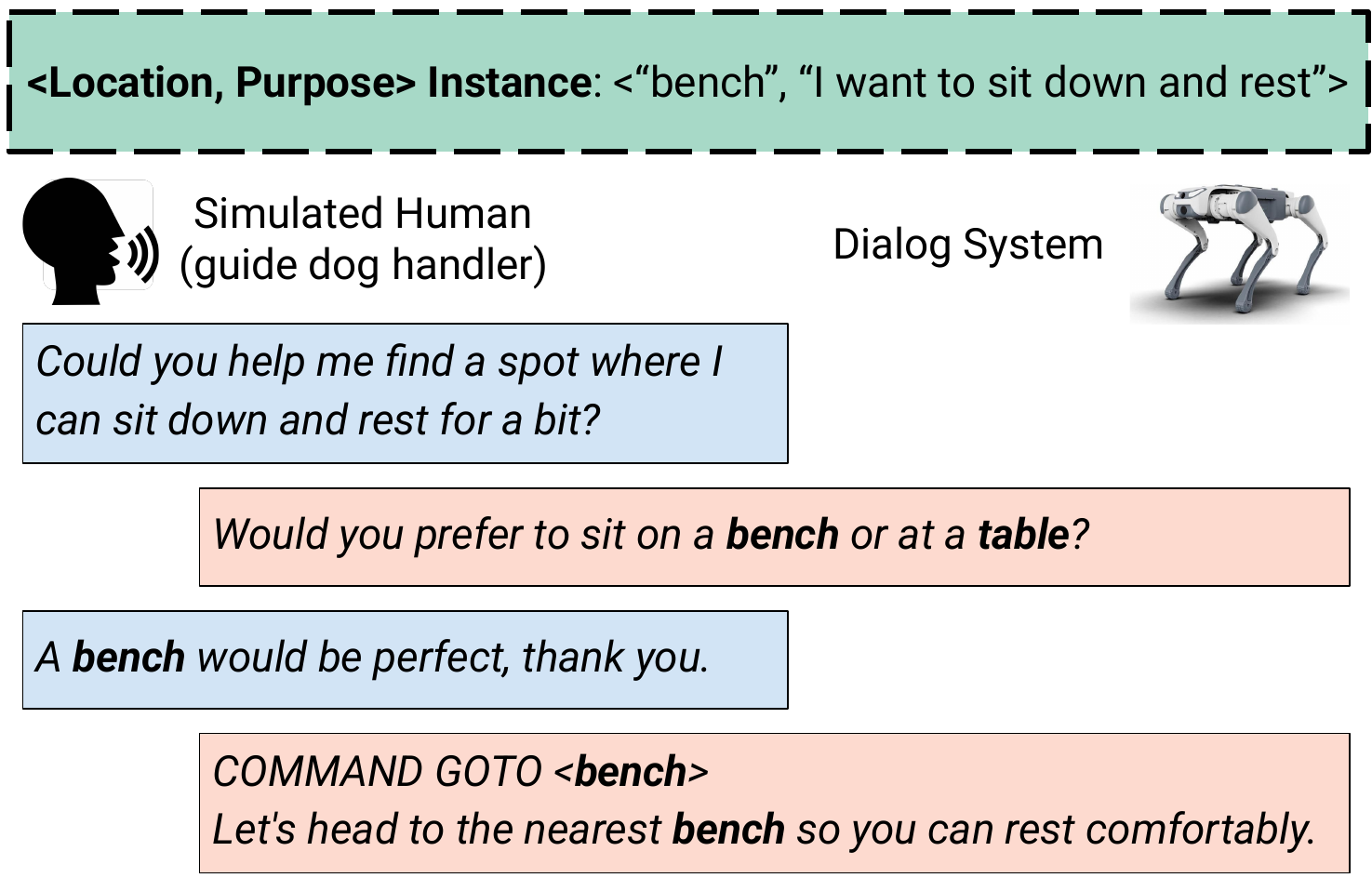}
    \caption{An example simulated conversation. The handler implicitly states a purpose sampled from \texttt{<location, purpose>} pair. The dialog system suggests relevant locations. After confirming the handler is looking for a bench, the system generates a formal task and concludes the task specification conversation.
    }
    \label{fig:sim_conv}
\end{center}
\end{figure}

\paragraph{Simulated Human} 

We used GPT-4 to simulate a visually impaired user, instructing with the following prompt:

\begin{quote}
    ``\emph{You are an AI chatbot pretending to be a visually impaired person. You need to navigate to \texttt{location}. Do not say the name of where you want to go, unless asked. You are only interested in going to \texttt{location} and must not show interest in any other locations.}''
\end{quote}

This prompt yields ambiguity by discouraging the direct mention of the location. 
Each conversation was initiated by providing the simulated human with the \texttt{purpose} from a \texttt{<location, purpose>} pair (e.g., \textit{``Begin the conversation by indicating that you want something to drink''}). 
An example dialog is shown in Figure~\ref{fig:sim_conv}.

\subsection{Evaluation of Accuracy and Efficiency}

To test \textbf{Hypothesis I}, we simulated conversations for all 77 task pairs and compared three approaches:
\begin{enumerate}
    \item \textbf{Keyword-Based}: A multi-turn template-based dialog system using keyword detection. If no keywords are found, it lists all known locations.
    \item \textbf{Single-Turn}: Our LLM-based system, but restricted to a single turn (no clarification questions).
    \item \textbf{Ours}: Our full multi-turn LLM-based dialog system, described in Section~\ref{sec:methodology}.
\end{enumerate}

For multi-turn approaches, dialogs were limited to six turns; exceeding this limit counted as a failure. 
A trial was successful if the dialog concluded with the correct (or functionally equivalent) formal task. 
We measured accuracy and efficiency by the dialog length and its success rate.



\begin{figure}[t]
\begin{center}
    \includegraphics[scale=0.25]{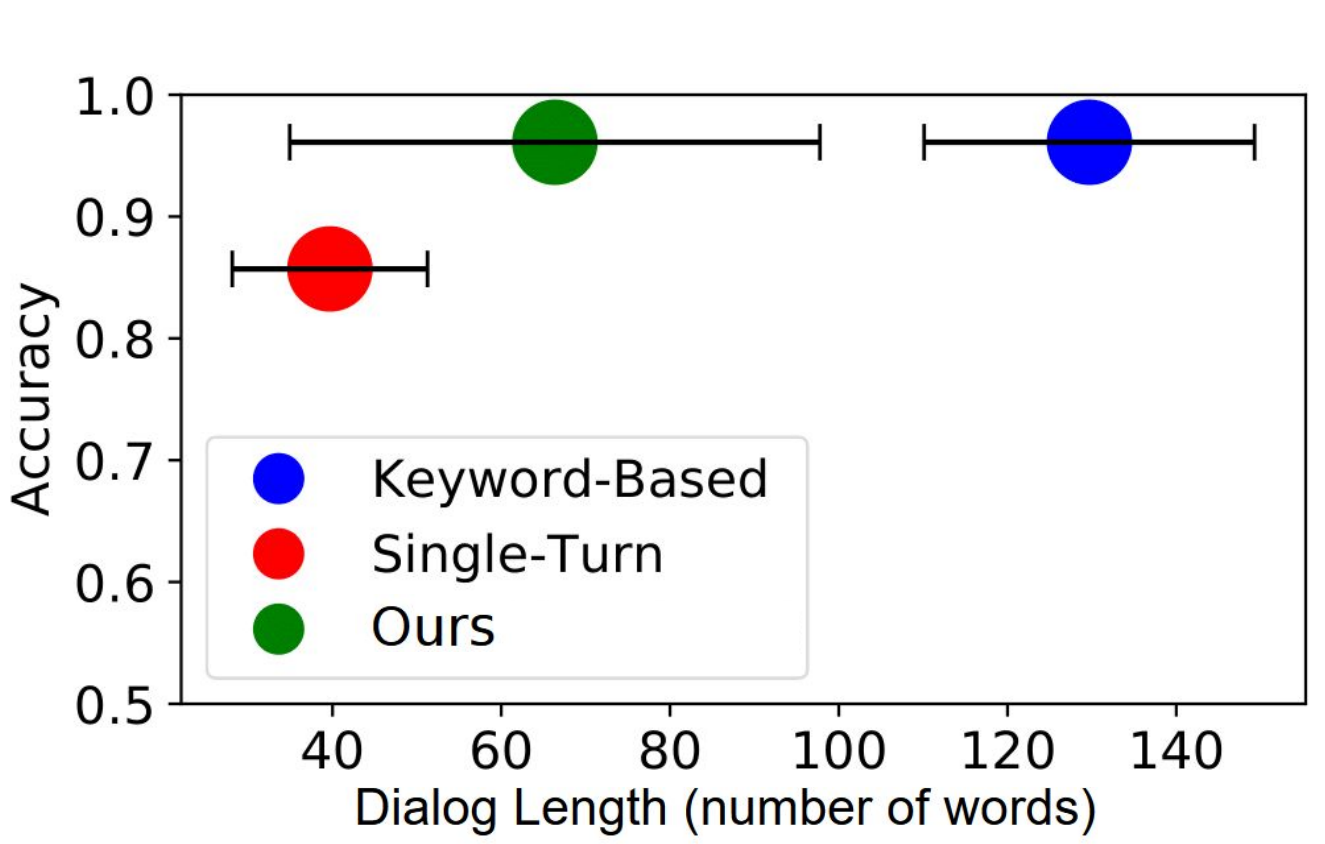}
    \caption{Accuracy vs. Average Dialog Length without noise. Our approach provides the best trade-off between high accuracy and efficient conversation.}
    \label{fig:scatter_plot_unperturbed}
\end{center}
\end{figure}

As shown in Figure~\ref{fig:scatter_plot_unperturbed}, the result highlights the trade-offs between each approach in terms of accuracy, average number of dialog turns. 
It achieves the highest accuracy (94.8\%) while being more concise than the Keyword-Based baseline, which is overly verbose as it often must enumerate all locations. 
These results support \textbf{Hypothesis I}.


\subsection{Evaluation of Robustness to Noisy Speech}


To test robustness to speech recognition errors that may arise from speech-to-text models (\textbf{Hypothesis II}), we simulated noisy input by perturbing each character in the user's responses with $0.3$ probability, randomly applying a deletion, insertion, or substitution.
This is particularly relevant to a guide dog setting, in which the visually impaired person can navigate in crowded and noisy environments. 

\begin{figure}[ht]
\begin{center}
    \includegraphics[scale=0.25]{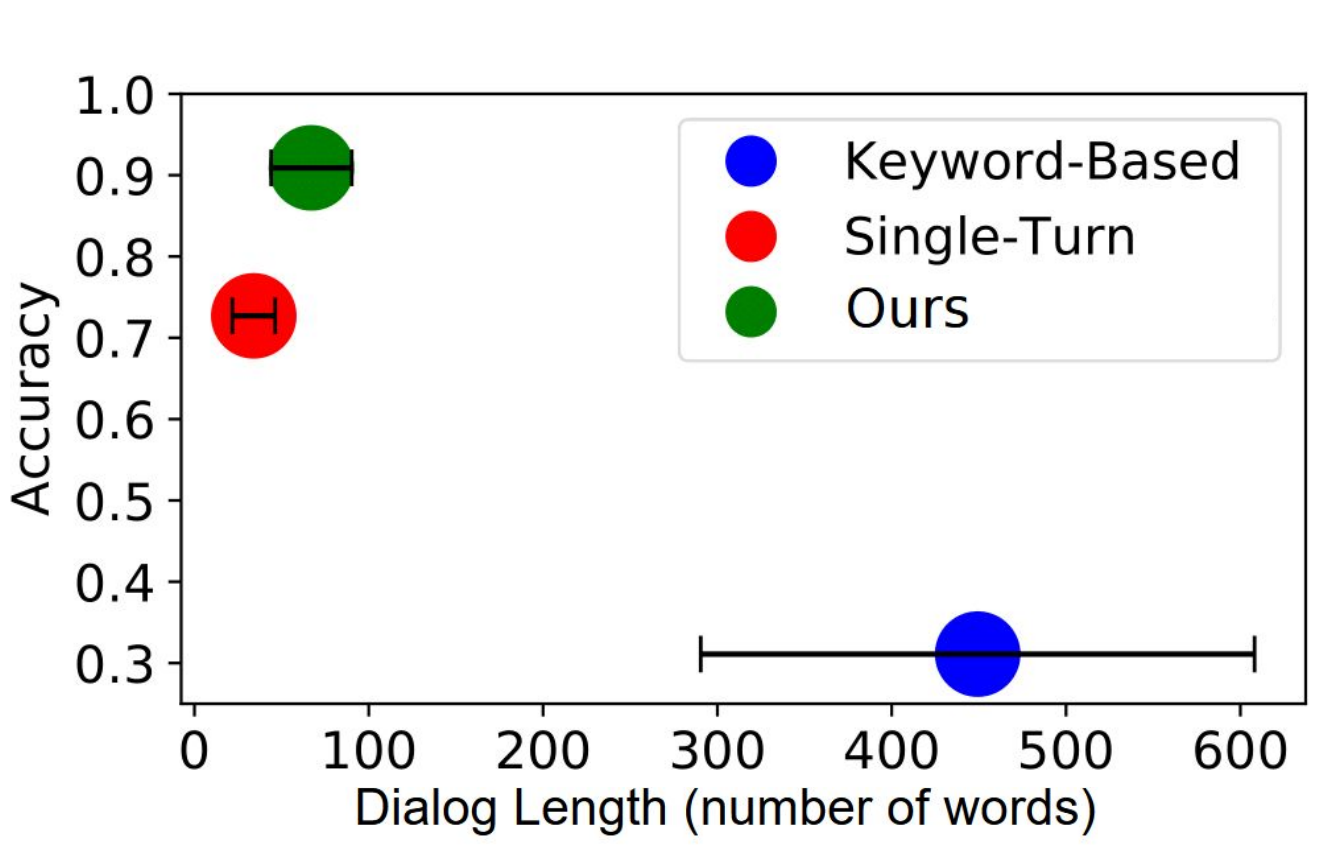}
    \caption{Accuracy vs. Dialog Length with perturbed (noisy) inputs. Our approach remains highly accurate, while the keyword-based method fails.}
    \label{fig:scatter_plot_perturbed}
\end{center}
\end{figure}

Figure~\ref{fig:scatter_plot_perturbed} shows the results that suggest our dialog system is robust to perturbations, only losing 5.2\% accuracy, and marginally increasing dialog length.
In contrast, the Keyword-Based's performance collapsed, as it failed to match the perturbed keywords. 
This supports \textbf{Hypothesis II}.

\begin{table}[ht]
\small
    \centering
    \begin{tabular}{ |c|c| } 
     \hline
     Location & Purpose \\ 
     \hline
      elevator & ``I want to go to the 2nd floor''  \\ 
     \hline
     water fountain & ``I want something to drink'' \\ 
     \hline
     waiting room & ``I want to sit down'' \\ 
     \hline
    \end{tabular}
    \caption{The three location-purpose pairs used for planner-informed evaluation.}
    \label{tab:planner_pairs}
\end{table}

\subsection{Evaluation of Planner-Informed Dialog}

Finally, we evaluated if providing plan information (e.g., travel distance) in the dialog helps the user choose more efficient routes (\textbf{Hypothesis III}). 
We used three tasks (Table~\ref{tab:planner_pairs}) where multiple functionally equivalent destinations existed at different distances.

\begin{table}[ht!]
\small
    \centering
    \begin{tabularx}{\linewidth}{ |c|X|X|X| } 
     \hline
     Method & Dialog Len. & Nav. Cost & Total Time \\ 
     \hline
     \textbf{w/ Plan Info} & 115.67 & \textbf{55.17} & \textbf{230.15} \\ 
     \hline
     \textbf{w/o Plan Info} & \textbf{84.83} & 73.00 & 277.27 \\      
     \hline
    \end{tabularx}
    \caption{Effect of Plan Information on Performance Metrics}
    \label{tab:planner_sim}
\end{table}

We compared our system with and without plan information. Table~\ref{tab:planner_sim} shows that when the dialog included navigation costs, the simulated user consistently chose the closer destination. 
Although this increased dialog length, the reduction in navigation cost led to a lower overall task time (calculated assuming 2.5 words/sec talking and 0.3 m/s walking speeds). 
These results support \textbf{Hypothesis III}.


\section{Conclusion}
\label{sec:conclusion}

In this paper, we developed a novel robotic guide dog system that can verbally convey both navigation plans and environmental scenes using a task planner and an LLM.
The effectiveness was validated through two evaluations.
The human study with visually impaired participants confirmed that combined plan and scene verbalization is an effective and preferred communication strategy.
The simulation analysis supported this, demonstrating high accuracy, robustness, and efficiency by leveraging planner knowledge, which accounts for the high user satisfaction.

Despite these promising results, the human study highlighted key challenges, particularly a lower safety rating compared to biological counterparts, underscoring the need for further enhancements.
Future work will involve expanded user studies, increased system autonomy, and deployment in more complex environments.
The ultimate goal is to integrate robotic guide dogs into daily life, enhancing independent navigation for both visually impaired and sighted individuals.


\section*{Ethical Statement}

The research protocol involving human participants was reviewed and approved by the Institutional Review Board (IRB). 
Seven legally blind individuals were recruited for the study. All participants provided written informed consent after being fully briefed on the experimental task and were compensated for their time. 
Participants were explicitly informed that their participation was voluntary and that they retained the right to withdraw from the study at any time without penalty.
To ensure participant safety during the real-world navigation component, a Wizard-of-Oz approach was used. 
In this setup, an expert operator remotely controlled all physical movements of the robot. This protocol was implemented to mitigate risks and isolate the evaluation to the system's dialog strategies. 
Researchers provided support by reading instructions and questionnaire items to participants.
Data was stored securely in accordance with institutional data protection policies.
This work contributes to the field of assistive robotics, which focuses on the well-being of people with disabilities. 
The primary objective of this research is to enhance the independence and mobility of visually impaired individuals by addressing the limitations posed by the scarcity of guide dogs and the challenges of their care.

\section*{Acknowledgments}
The authors thank Thomas Panek, Bill Ma, and Ken Fernald for their guidance in this research, and the participants for their valuable feedback. A portion of this work has taken place at the Autonomous Intelligent Robotics (AIR) Group, SUNY Binghamton. AIR research is supported in part by the NSF (IIS-2428998, NRI-1925044), Ford Motor Company, DEEP Robotics, OPPO, Guiding Eyes for the Blind, and SUNY RF.  Research reported in this publication was supported in part by SUNY System Administration using the SUNY AI Platform. 

\bibliography{aaai2026}

\end{document}